\documentclass{styles/svproc}
\usepackage{url}
\usepackage{cite}
\usepackage{amsmath,amssymb,amsfonts}
\usepackage{algorithmic}
\usepackage{graphicx}
\usepackage{textcomp}
\usepackage{xcolor}
\usepackage{algorithm,algorithmic}

\makeatletter
\newcommand{\printfnsymbol}[1]{%
  \textsuperscript{\@fnsymbol{#1}}%
}
\makeatother

\begin{document}
\mainmatter              
\title{Retinal Fundus Multi-Disease Image Classification using Hybrid CNN-Transformer-Ensemble Architectures}

\author{
    Deependra Singh\textsuperscript{1}\thanks{The first two authors contributed equally to this work and should be regarded as Joint First Authors},  
    Saksham Agarwal\textsuperscript{2}\printfnsymbol{1},  
    Subhankar Mishra\textsuperscript{3,\$}
}

\institute{
    \textsuperscript{1}School of Physical Science,  
    \textsuperscript{2}School of Biological Science,  
    \textsuperscript{3}School of Computer Science, 
    National Institute of Science Education and Research, Bhubaneswar, India \\  
    \email{\{\textsuperscript{1}deependra.singh, \textsuperscript{2}saksham.agarwal, \textsuperscript{3}smishra\}@niser.ac.in} \\  
    \textsuperscript{\$} Corresponding author: Subhankar Mishra, \email{smishra@niser.ac.in}
}


\maketitle       

\begin{abstract}
Our research is motivated by the urgent global issue of a large population affected by retinal diseases, which are evenly distributed but underserved by specialized medical expertise, particularly in non-urban areas. Our primary objective is to bridge this healthcare gap by developing a comprehensive diagnostic system capable of accurately predicting retinal diseases solely from fundus images. However, we faced significant challenges due to limited, diverse datasets and imbalanced class distributions. To overcome these issues, we have devised innovative strategies. Our research introduces novel approaches, utilizing hybrid models combining deeper Convolutional Neural Networks (CNNs), Transformer encoders, and ensemble architectures sequentially and in parallel to classify retinal fundus images into 20 disease labels. Our overarching goal is to assess these advanced models' potential in practical applications, with a strong focus on enhancing retinal disease diagnosis accuracy across a broader spectrum of conditions. Importantly, our efforts have surpassed baseline model results, with the C-Tran ensemble model emerging as the leader, achieving a remarkable model score of 0.9166, surpassing the baseline score of 0.9. Additionally, experiments with the IEViT model showcased equally promising outcomes with improved computational efficiency. We've also demonstrated the effectiveness of dynamic patch extraction and the integration of domain knowledge in computer vision tasks. In summary, our research strives to contribute significantly to retinal disease diagnosis, addressing the critical need for accessible healthcare solutions in underserved regions while aiming for comprehensive and accurate disease prediction.
\end{abstract}

\section{Introduction }
The human retina serves a critical role in the visual system by converting incoming light signals into electrical/chemical signals that are subsequently processed by the brain. Unfortunately, there are numerous retinal diseases that can lead to irreversible damage, resulting in permanent vision loss. Timely diagnosis and treatment are therefore essential to manage and prevent further deterioration. However, the availability of quality eye care can vary significantly within countries and regions, leading to limited access, particularly for individuals residing in remote areas \cite{b1}.

Addressing this issue is crucial to ensure equitable access to quality eye care. The increasing prevalence of specific retinal diseases emphasizes the importance of early detection and efficient diagnostic systems. For instance, glaucoma is predicted to impact 111.8 million individuals by 2040 \cite{b2} predicted. \cite{b3} highlights that out of the 415 million people worldwide living with diabetes in 2015, 145 million experienced diabetic retinopathy (DR). Approximately 6.2 million individuals globally are affected by age-related macular degeneration (AMD) \cite{b4}. Various retinal diseases, including age-related macular degeneration (AMD), exhibit distinct pathophysiological features that can serve as valuable sources for the development of artificial intelligence (AI) diagnostic tools. Pathologies such as drusen or choroidal neovascularization (CNV), subretinal hemorrhage, and vascular leakage in dry or neovascular AMD provide potential photographic markers for the development of AI-based diagnostic algorithms \cite{b5}. These markers and other diagnostic indicators enable ophthalmologists to identify and differentiate between different retinal conditions \cite{b6}. The development of image-based diagnostic systems holds significant promise for addressing these challenges. \\

In recent years, there has been a growing interest in AI-based diagnostic systems that utilize fundus images.  Initially, these systems targeted specific retinal diseases, such as diabetic retinopathy \cite{b7} and AMD \cite{b8}. However, \ a major drawback of these approaches is their narrow focus on individual diseases or disorders. Patients often experience multiple retinal diseases in real-world scenarios, rendering these software solutions less effective \cite{b9}.To address this challenge, there has been a shift towards developing multi-disease classification systems for retinal diseases \cite{b10}. Prior studies have investigated deep-learning methods, mostly Convolutional neural networks (CNNs), both standalone \cite{b11}\cite{b12}, or ensembled \cite{b13}\cite{b14}, to diagnose retinal diseases using fundus images. Notably, \cite{b10} used CNNs to classify retinal diseases with impressive accuracies. \cite{b11} explored transfer learning, achieving $93.58\%$ accuracy. Ensemble approaches, like \cite{b13}, attained an AUC score of 0.97, even for limited labels. While CNNs have been effective, recent advancements in medical diagnosis have involved transformer-based models in increasing accuracy and targeting more disease labels \cite{b15} \cite{b16}. However, challenges persist in achieving better classification accuracy, enhancing model generalization and robustness, and deploying models in resource-limited environments; knowing the availability of large, properly labeled, and balanced datasets is scarce \cite{b11}. 

These works showcase deep learning's impact, though dataset limitations and lower accuracy on certain labels warrant attention. Our research seeks to pioneer a hybrid model that addresses these challenges, providing a scalable solution for comprehensive retinal disease diagnosis."

\subsection{Contribution}
 To build our hybrid models combining CNN and transformers, we selected two baseline models from previous works: C-Tran \cite{b15}, which was previously been tested on a fundus image dataset with 20 labels, and IEViT \cite{b16}, which was previously been tried on chest X-ray images. Our work has the following contribution to current research:
\begin{enumerate}
    \item  We developed six novel models that combine deeper CNN backbones, transformer encoders, and ensemble architectures.
     \item All six variants were tested on the dataset, showing improved performances over the baseline models.
    \item Utilisation of the domain knowledge to highlight more relevant image regions with importance patch extraction. 
    \item Improvement in the general transformer-based learning using the IE concept.
    \item For the purpose of reproducibility, the source code files can be accessed on the GitHub Repository.\footnote{\;https://github.com/smlab-niser/23retinald}
\end{enumerate} 

\section{Dataset}
Among the three publicly available datasets shown in the table below, we chose the MuReD dataset \cite{b17}. This decision was influenced by its utilization in the C-Tran paper we aimed to reproduce for establishing baselines of our model \cite{b15}. MuReD, a refined variant of the highly imbalanced RFMiD dataset, was preferred due to its relevance. In the future, we intend to incorporate the RFMiD 2.0 and JSIEC datasets in our analysis as it is the latest dataset out of three and encompasses all the essential labels of MuReD. \\

\begin{table}[H]
\centering
\caption{List of Publicly available datasets}
\begin{tabular}{|l|l|l|l|l|l|}
\hline
\textbf{Dataset Name} & \textbf{Train} & \textbf{Validation} & \textbf{Test} & \textbf{Total} & \textbf{Labels} \\
\hline
RFMiD dataset \cite{b18} & 1920 & 640 & 640 & 3200 & 46 \\
MuReD dataset \cite{b17} & 1766 & 442 & - & 2208 & 20\\
RFMiD 2.0 \cite{b19} & \;\;516 & 344 & 344 & \;\;860 & 49\\
JSIEC \cite{b12} & - & - & - & 1000 & 39 \\
\hline
\end{tabular}
\label{tab1}
\end{table} 

\begin{table}[H]
\centering
\caption{List of Diseases to be detected \cite{b15}}
\begin{tabular}{|c|c|c|c|c|}
\hline
Acronym & Full Name & Training & Validation & Total \\
\hline
DR & Diabetic Retinopathy & 396 & 99 & 495 \\
NORMAL & Normal Retina & 395 & 98 & 493 \\
MH & Media Haze & 135 & 34 & 169 \\
ODC & Optic Disc Cupping & 211 & 52 & 263 \\
TSLN & Tessellation & 125 & 31 & 156 \\
ARMD & Age-Related Macular Degeneration & 126 & 32 & 158 \\
DN & Drusen & 130 & 32 & 162 \\
MYA & Myopia & 71 & 18 & 89 \\
BRVO & Branch Retinal Vein Occlusion & 63 & 16 & 79 \\
ODP & Optic Disc Pallor & 50 & 12 & 62 \\
CRVO & Central Retinal Vein Oclussion & 44 & 11 & 55 \\
CNV & Choroidal Neovascularization & 48 & 12 & 60 \\
RS & Retinitis & 47 & 11 & 58 \\
ODE & Optic Disc Edema & 46 & 11 & 57 \\
LS & Laser Scars & 37 & 9 & 46 \\
CSR & Central Serous Retinopathy & 29 & 7 & 36 \\
HTR & Hypertensive Retinopathy & 28 & 7 & 35 \\
ASR & Arteriosclerotic Retinopathy & 26 & 7 & 33 \\
CRS & Chorioretinitis & 24 & 6 & 30 \\
OTHER & Other Diseases & 209 & 52 & 261 \\
\hline
\end{tabular}
\end{table}

\section{Preprocessing of images}

\subsection{Sampling Techniques}
Several sampling techniques have been employed in the literature, including LP ROS \cite{b15}, Dynamic Random Sampling \cite{b12}, and Weighted Random Sampler. We focused on two specific sampling techniques, Weighted Random Sampling and LP-ROS (Label Powerset Random Oversampling), listed in table \ref{tab2}. While the CTran paper used LP ROS for addressing the class imbalance, the Weighted Random Sampling technique, implemented in the code provided, demonstrated better performance. The decision to utilize Weighted Random Sampling was based on empirical evidence suggesting its efficacy in handling class imbalance within the CTran model.

\begin{table}[H]
\centering
\caption{Sampling techniques used}
\begin{tabular}{|c|p{9cm}|}
\hline
\centering \textbf{Sampling Technique} &  \textbf{Description} \\ \hline
\vspace{0.05cm}Weighted Random Sampling & Addresses class imbalance by assigning higher weights to underrepresented classes during training. \\
\hline
\centering LP-ROS (10\%) & An oversampling technique based on the Label Powerset transformation, used to handle class imbalance. Here, the minority class samples are augmented by adding $10\%$ more samples to balance the class distribution. \\
\hline
\end{tabular}
\label{tab2}
\end{table}

\subsection{Augmentation}
In the context of image data preprocessing, augmentation techniques are commonly used to increase the diversity and variability of the training dataset. These techniques introduce modifications or transformations to the original images, allowing the model to learn from a wider range of data patterns. These augmentation techniques are applied as part of the data transformation pipeline using the torchvision.transforms module. They contribute to enhancing the model's robustness and ability to generalize patterns from the training data. The following table summarizes the augmentation techniques applied in this work:

\begin{table}[H]
\centering
\begin{tabular}{|c|p{8cm}|}
\hline
\textbf{Technique} & \textbf{Description} \\ \hline
Random Horizontal Flip & Randomly flips the image horizontally with a probability of 0.5. \\ \hline
Random Vertical Flip & Randomly flips the image vertically with a probability of 0.5. \\ \hline
Random Rotation & Randomly rotates the image by 15 degrees. \\ \hline
Colour Jitter & Adjusts the brightness, contrast, saturation, and hue of the image with values of 0.4, 0.4, 0.4, and 0.1 respectively. \\ \hline
Resize & Resizes the image to the specified size. \\ \hline
ToTensor & Converts the image to a tensor. \\ \hline
Normalize & Normalizes the image tensor using pre-defined mean and standard deviation values of [0.485, 0.456, 0.406] and [0.229, 0.224, 0.225], respectively. \\ \hline
\end{tabular}
\end{table}

\section{Models}
This section provides an overview of the models used in our study and highlights their unique contributions and variations.
\subsection{Deep Convolutional Neural Networks}
We explored various deep CNN architectures as backbone models to identify the most effective model configuration. Our selection encompassed ResNet152d, DenseNet121, DenseNet201, and EfficientNetV2Small. These models were pre-trained on ImageNet and served as feature extractors, generating feature embeddings that were subsequently processed through transformer layers for classification. By evaluating different CNN architectures, we sought to capitalize on their unique feature extraction capabilities. Additionally, we utilized denser versions of DenseNet and ResNet compared to those employed in \cite{b15} (DenseNet121), aiming to enhance the performance of the baseline model. The performance of these architectures on our dataset is given in table \ref{tab4}.

\begin{table}[H]
\centering
\caption{Performance of each CNN architecture on the dataset}
\begin{tabular}{|c|c|c|c|}
\hline
\textbf{Model} & \textbf{Loss} & \textbf{ML F1} & \textbf{ML AUC} \\
\hline
DenseNet121 & 2.121  & 0.276 & 0.951  \\\hline
\textbf{DenseNet201}$^{\mathrm{a}}$ & 1.692  & 0.272 & 0.959  \\\hline
\textbf{ResNet152d} & 1.815  & 0.410 & 0.947  \\\hline
EfficientNetV2-S & 1.881  & 0.214 & 0.958  \\\hline
\multicolumn{4}{c}{$^{\mathrm{a}}$ CNNs in bold indicate the top two best-performing architectures} 
\end{tabular}%
\label{tab4}
\end{table}

\subsection{Classification Transformers C-Tran}

The C-Tran model employed in this project combines DenseNet201 and ResNet152d backbones with a Transformer encoder, facilitating good performance for multi-label classification. The key components of the C-Tran model encompass feature extraction, positional encoding, and label prediction.\\

For each input image, visual features are extracted using the chosen CNN backbone, resulting in a feature embedding $\textbf{X}_{\text{visual}}$ of dimensions $n \times b \times d$, where $n$ is the number of classes, $b$ batch size, and $d$ is the embedding dimension. To incorporate positional information, the model utilizes positional encoding, denoted as $\textbf{PE}_{2D}$, calculated using sine and cosine functions based on position in the height and width dimensions:
\begin{equation}
\textbf{PE}_{2D}(i, j, k) = 
\begin{cases}
\sin\left(\frac{i}{10000^{2k/d}}\right) & \text{if } k \text{ is even} \\
\cos\left(\frac{j}{10000^{2k/d}}\right) & \text{if } k \text{ is odd}
\end{cases}
\end{equation}
where $i, j$ represents the position in the height and width dimensions, $k$ represents the dimension of the positional encoding, and $d$ is the embedding dimension of the visual features.\\

The 2D positional encoding matrix, $\textbf{PE}_{2D}$, is element-wise added to the visual features after reducing it linearly to the dimensions of $\textbf{X}_{\text{visual}}$, enriching spatial awareness:
\begin{equation}
\textbf{X}_{\text{encoded}} = \textbf{X}_{\text{visual}} + \textbf{PE}_{2D}
\end{equation}
where $\textbf{X}_{\text{encoded}}$ represents the spatially enhanced visual features.\\

The Transformer encoder captures feature-label interactions using a self-attention mechanism, fostering the learning of diverse dependencies between features and labels. The final label predictions, $\textbf{Y}_{\text{pred}}$, are generated using independent feed-forward networks for each label embedding:
\begin{equation}
\textbf{Y}_{\text{pred}} = \sigma(\textbf{E}_{\text{label}} \cdot \textbf{W} + \textbf{b})
\end{equation}
where $\sigma$ represents the sigmoid activation function, $\textbf{W}$ represents learned weights, and $\textbf{b}$ represents biases. The C-Tran algorithm synergizes feature extraction, positional encoding, and self-attention mechanisms within a Transformer architecture to achieve state-of-the-art multi-label classification performance.

\subsection{Ensemble Model}

The ensemble model combines the predictions from two distinct backbone models, DenseNet201 and ResNet152D, utilizing a weighted ensemble strategy. The ensemble model is configured with two variants: one that employs separate transformer layers for each backbone model and another that utilizes a single transformer layer for the combined embeddings from both.

\subsubsection{Variant 1: Separate Transformer Layers:}

In this variant, the input image undergoes two separate C-Tran paths. First, it passes through the DenseNet201 backbone model, generating feature embeddings. These embeddings then traverse through a set of transformer layers, as in C-Tran. The sigmoid function is applied to the output to produce the raw predictions denoted as $\textbf{P}_{\text{dn}}$. Simultaneously, the input image also traverses the ResNet152d backbone model, producing feature embeddings that pass through their respective transformer layers, and raw predictions are generated via sigmoid denoted as $\textbf{P}_{\text{rn}}$.\\

We employ weighted ratios, denoted as $a$ and $b$, to combine the predictions from both backbone models. These ratios determine the relative importance of each backbone model's predictions. The combined output embeddings, $\textbf{P}_{\text{c}}$, are calculated as follows:
\begin{equation}
\textbf{P}_{\text{c}} = a \cdot \textbf{P}_{\text{dn}} + b \cdot \textbf{P}_{\text{rn}}
\end{equation}
To form the combined loss $L$, the weighted average of these individual losses is computed:
\begin{equation}
L = a \cdot L_{\text{dn}} + b \cdot L_{\text{rn}}
\end{equation}

This combined loss, $L$, is then used for gradient descent optimization and model parameter updates. Additionally, evaluation metrics are computed using the sigmoid-activated predictions to assess model performance.

\subsubsection{Variant 2: Single Transformer Layer:}

In this variant, the input image is passed through both the DenseNet201 and ResNet152D backbone models. The feature embeddings from both backbones are then combined and fed into a single transformer layer. The output embeddings from both backbone models, denoted as $\textbf{X}_{\text{dn}}$ and $\textbf{X}_{\text{rn}}$, are concatenated along the $n$ dimension (dimension 0), creating a unified feature embedding, $\textbf{X}_{\text{c}}$: 
\begin{equation}
\textbf{X}_{\text{c}} = [\textbf{X}_{\text{dn}}, \textbf{X}_{\text{rn}}]
\end{equation}
where ${\scriptstyle[\; ]}$ operation signifies concatenation. This combined feature embedding, $\textbf{X}_{\text{c}}$, is subsequently processed by a single C-Tran transformer encoder layers, capturing the contextual information from both backbone models. The remaining steps, such as positional encoding, projection layer, and classification layer, remain the same as in the previous variant. This variant allows the model to leverage the strengths of both backbone models in a unified transformer layer.\\

\begin{algorithm}[H]
\caption{C-Tran Ensemble Variants Forward Pass}
\label{alg:ctrancoder1}
\begin{algorithmic}[1]

\REQUIRE{$x$: Input tensor of shape [batch, channel, height, width]}
\ENSURE{$p$: Output tensor of shape [batch, num\_classes]}

\STATE Initialize network components
\STATE $x_1 \gets \text{Backbone1}(x)$
\STATE $x_2 \gets \text{Backbone2}(x)$

\IF{Variant 1}
    \STATE Add positional encodings: $x_1, x_2 \gets (x_1 + P_1, \;x_2 + P_2)$
    \STATE $p_1 \gets \text{MLP}_1(\text{Transformer}_1(x_1))$
    \STATE $p_2 \gets \text{MLP}_2(\text{Transformer}_2(x_2))$
    \STATE $p = a\cdot p_1 + b \cdot p_2$
\ELSE
    \STATE $x \gets \text{Concatenate}(x_1, x_2)$
    \STATE Add positional encoding $P:$ $x \gets x + P$
    \STATE $p \gets \text{MLP}(\text{Transformer}(x))$
\ENDIF
\STATE Train model and update weights
\RETURN $p$
\end{algorithmic}
\end{algorithm}

\begin{figure}[H]
    \centering
    \includegraphics[height=6cm, width=10cm]{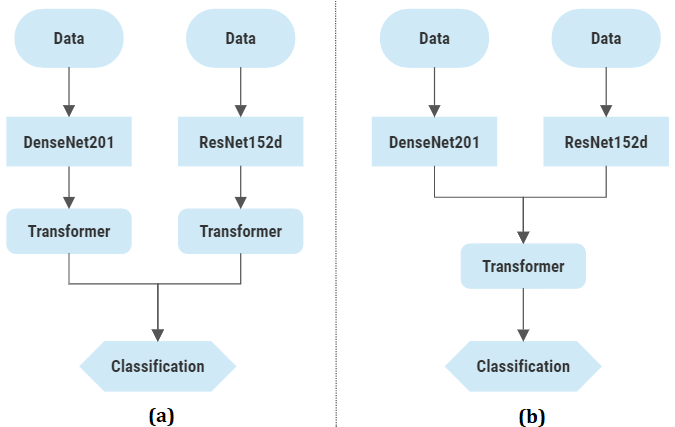}
    \caption{Models Representation for (a) Variant 1, (b) Variant 2 }
    \label{fig1}
\end{figure}

These two variants provide different approaches for integrating multiple backbone models and transformer layers within the ensemble model, enabling the model to leverage diverse features and capture complex patterns in the input data.

\subsection{IEViT Model}
The proposed IEViT (Iterative Expansion Vision Transformer) model is an extension of the ViT (Vision Transformer) architecture, incorporating concepts inspired by the ResNet (Residual Network). It introduces an iterative process in which the original input image is iteratively added to the output of each Transformer encoder layer. To facilitate this, a convolutional block is designed in parallel with the ViT network. The CNN block processes the entire input image and generates an embedding of the image denoted as $\mathbf{x}_{img}$. This embedding is then concatenated to the output of each Transformer encoder layer, denoted as $\mathbf{z}_l$, thereby incorporating the complete image information throughout the encoding process.\\

The IEViT architecture comprises stacked 2D convolutional layers, followed by a 1D global maximum pooling layer within the CNN block. The resulting tensor is a feature embedding, $\mathbf{x}_{\text{img}}$ of dimension D. Additionally, the image is divided into $N = \frac{H \cdot W}{P^2}$ patches, which are flattened to patches $x_p$ of size $N \times (P^2 \cdot C)$. Subsequently, the patch embeddings $z_l$ are created by mapping the patches to D dimensions using 2D convolution layers. Position embeddings denoted as $\mathbf{E}_{\text{pos}}, {E}_{\text{pos}} \in \mathbb{R}^{(N+1) \times D}$, are introduced to encode the positional information of each patch in the original image.\\

Firstly, the initial patch embedding $\mathbf{z}_{\text{p}}$ and the positional encodings $\mathbf{E}_{\text{pos}}$ are element-wise added. An optional class token might also be concatenated with $z_p$.

\begin{equation}
\mathbf{z}_{\text{0}} = \mathbf{z}_{\text{p}} + \mathbf{E}_{\text{pos}}
\end{equation}

where $\mathbf{z}_{\text{0}}$ represents the combined embeddings. The core iterative expansion process in IEViT occurs during the transformation of image embeddings through stacked Transformer encoder layers. At each layer $l$, the image embedding, $\mathbf{x}_{\text{img}}$, is fused with the encoder output, $\mathbf{z}_l$, resulting in a combined representation denoted as $\mathbf{z}^{\hat{l}}$. This process is mathematically represented as:

\begin{equation}
\mathbf{z}^{\hat{l}} = [\mathbf{z}_l, \mathbf{x}_{\text{img}}], \quad \text{where } \mathbf{z}^{\hat{l}} \in \mathbb{R}^{(N+1+l) \times D}.
\end{equation}

The iterative expansion in IEViT enhances its capability to capture spatial information and maintain global context, contributing to superior performance across diverse computer vision tasks.\\

We propose two novel variants to enhance the patch extraction step in IEViT, leveraging domain knowledge of the 20 diseases targeted for classification \cite{b16}. These variants address the importance of prioritizing patches from near the center of the image, where crucial features are expected to be prominent for the specific disease labels we're targeting.\\

\subsubsection{Variant 1- Unequal Patches:}
In this variant, we manually set the patch dimensions to be extracted from the image for forming the patch of embedding of dimension D. We extract patches with decreasing dimensions as we move closer to the center. This approach allocates more space to patches near the center, where crucial features are expected to be prominent. The patch dimensions, manually defined as 32x32, 16x16, and 8x8, progressively decrease towards the center.\\

\subsubsection{Variant 2- Dynamic Patch Decomposer (DPD):}

This approach utilizes normalized learnable parameter importance weights for the initial 144 patches, each with a dimension of $32\times32$. The method dynamically determines which patches to subdivide based on their importance, using the formula for the new patch dimension (NPD):
\[NPD = 32 \times \left(\frac{{w_{avg}}}{{w_i}}\right)^k\] 
where $w_i$ represents the weight of the $i$-th patch, $w_{avg}$ is the average weight, and k is a hyperparameter. Patches with $NPD>32$ remain unchanged, while others are subdivided into dimensions $NPD\times NPD$ after rounding NPD to the nearest factor of 32. The new patches are extracted, and their embedding $z_n$ is appended to the former patch embedding tensor $z_l$ at the position of the original patches that were divided or merged, as represented by the equation:

\[z_l'= z_l \, \oplus \, z_n\]

A similar adjustment is applied to maintain positional encoding and adapt it for subdivided or merged patches. Specifically, the positional encoding is extended to accommodate new patches while preserving the existing ones, ensuring that spatial information is accurately represented throughout the encoding process. The positional encoding for the new patches is inserted at the position of the original patch in the encoding tensor. This extension is expressed by:

\[E_{pos}' = E_{pos} \, \oplus \, E_{new}\]

Additionally, the MLP (Multi-Layer Perceptron) head parameters are dynamically adjusted to handle the input changes due to patch subdivision and merging. Parameter sharing is employed to efficiently accommodate the new patches without a significant increase in model parameters, ensuring that the MLP weight matrix adapts to the varying input dimensions. Weight splicing is used when the patches are merged. This adaptive approach enhances the model's ability to capture and utilize original and subdivided patch information.

\begin{algorithm}
\caption{DPD Patch Division and Weight Handling}
\label{alg:ctrancoder2}
\begin{algorithmic}[1]
\STATE Calculate $NPD = 32 \times \left(\frac{{w_{avg}}}{{w_i}}\right)^k$
\FOR{patch in patch embedding $z_0$}
    \IF{$NPD > 32$}
        \STATE \textit{No patch subdivision required}
    \ELSE
        \STATE \textit{Subdivide the patch into $NPD\times NPD$ sized patches}
        \STATE Extract the embeddings for the new patches  \STATE $z_l'= z_l \, \oplus \, z_n$
        \STATE Adjust positional encoding: $E_{pos}' = E_{pos} \, \oplus \, E_{new}$
    \ENDIF
\ENDFOR
\IF{$\text{MLP input dimension increases}$}
    \STATE Expand MLP head parameters with weight sharing
\ELSIF{$\text{MLP input dimension decreases}$}
    \STATE Slice MLP head parameters
\ELSE
    \STATE MLP head parameters remain unchanged
\ENDIF 
\STATE Continue with standard processing
\end{algorithmic}
\end{algorithm}

\newpage
\subsection{CTran Variants with IEViT Concept}

In this subsection, we present two variants of the CTran model that incorporate the Iterative expansion concept from IEViT.

\subsubsection{IECTe- Iterative Expansion C-Tran Ensembler:}

The IECTe variant combines ensemble model variant 2 with the IE concept by appending the reduced feature embeddings from the two CNN backbones $x$ and $y$ to each output of transformer layers. Concatenating feature embeddings along the $num\_classes$ dimension (dim = 0):
\[z_0 = [x, y]\]
\[\mathbf{z}^{\hat{l}} = [\mathbf{z}_l, \mathbf{z_0}]\]
By incorporating these additional feature embeddings, the model gains a holistic understanding of the input image from multiple perspectives. This enables the model to leverage the strengths of both ResNet201 and DenseNet201 architectures, leading to enhanced representation learning and improved performance. The iterative nature of the IEViT approach ensures that the network retains knowledge of the entire input image throughout the Transformer encoding process. By iteratively concatenating the feature embeddings to the Transformer output, the IECTe model continuously "remembers" the full image information, facilitating effective information fusion and promoting robust representations.

\subsubsection{IeECT: Iterative Ensemble Expansion C-Tran-}
The IeECT variant focuses on reducing the overall model complexity for IECTe. In this variant, the feature embedding from DenseNet201 serves as the primary input to the Transformer $(z_0 = x)$, while the feature embedding from ResNet152 is appended to the Transformer output after each layer.
\[\mathbf{z}^{\hat{l}} = [\mathbf{z}_l, \mathbf{y}]\]
By employing this design, the IeECT model benefits from the expressive power of DenseNet201, which captures intricate image details and relationships. Simultaneously, the inclusion of the ResNet152 feature embedding allows the model to incorporate information from a different architectural perspective, enriching the representations with complementary insights. Using the IEViT concept in IeECT ensures that the model retains knowledge of the original image throughout the encoding process while maintaining fewer parameters.\cite{b1}
\vspace{-0.4cm}
\begin{algorithm}[H]
\caption{IE pipeline Variants for CTran}
\label{alg:ctrancoder3}
\begin{algorithmic}[1]

\STATE Initialize network components, L transformer encoder layers
\STATE $x_1, x_2 \gets \text{Backbone1}(x), \text{Backbone2}(x)$

\IF{Variant 1}
    \STATE $z_0, \;z \gets \text{Concatenate}(x_1, x_2)$
\ELSE
    \STATE $z_0,\; z \gets x_1, \;x_2$
\ENDIF
\STATE Add positional encoding $P:$ $z_0 \gets z_0 + P$
\STATE $z_1 \gets \text{Transformer\_Layer}_0(z_0)$
\FOR{layer l in transformer encoder $l\neq 0$}
    \STATE $z_l \gets \text{Concatenate}(z_l, z)$
    \STATE $z_(l+1) \gets \text{Transformer}(z_1)$
\ENDFOR
\RETURN $p \gets \text{MLP}(z_L)$
\end{algorithmic}
\end{algorithm}
\vspace{-1.2cm}
\begin{figure}[H]
    \centering
 \includegraphics[height=5.3cm, width=8cm]{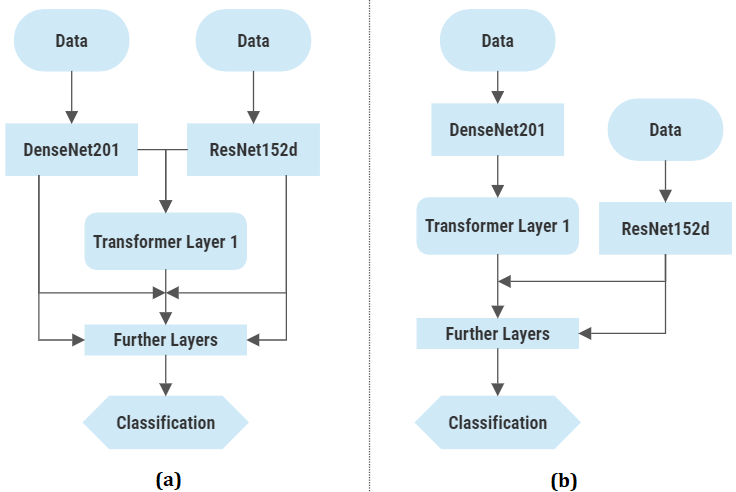}
    \caption{Model Representation for:(a) IECTe, (b) IeECT }
    \label{fig2}
\end{figure}
\vspace{-1cm}
\newpage
\section{Model Training}

\subsection{Hyperparameters}
\vspace{-0.5cm}
\begin{table}[H]
\caption{List of hyperparameters used for model optimization}
\centering
\begin{tabular}{|p{3cm}|p{7cm}|}
\hline
\textbf{Hyperparameter} & \textbf{Value} \\
\hline
Number of Epochs & 200 \\\hline
Batch Size & 8, 12, 16, 32 \\\hline
Embedding Dimensions & 960 \\\hline
Optimizer & AdamW \\\hline
Scheduler & Cosine Annealing Warm Restarts,
Learning Rate: 5e-5,
Weight Decay: 1e-6 \\\hline
Loss Function & BCEwithLogitLoss \\\hline
Transformer Layers & 6, 12 \\
\hline
\end{tabular}
\label{tab5}
\end{table}
\vspace{-0.7cm}
\subsection{Metrics}
To evaluate the performance of our models, we used the same metrics as used in \cite{b15} for comparison on the same dataset. These metrics include F1-score, AUC, and mAP. ML mAP, ML F1, and ML AUC are calculated by averaging scores for each label, excluding the "NORMAL" label- having the AUC and F1-score represented by Bin AUC and Bin F1 metric. 
\[{ \text{ML Score} = \frac{\text{ML mAP} + \text{ML AUC}}{2}}\]
The Model Score metric evaluates the overall performance by considering both disease classification and normalcy detection. 
\[\text{Model Score} = \frac{\text{ML Score} + \text{Bin AUC}}{2}\]
 \vspace{-0.5cm}

\section{Results and Discussion}

\begin{figure}[H]
    \centering
    \includegraphics[height=6cm, width=9cm]{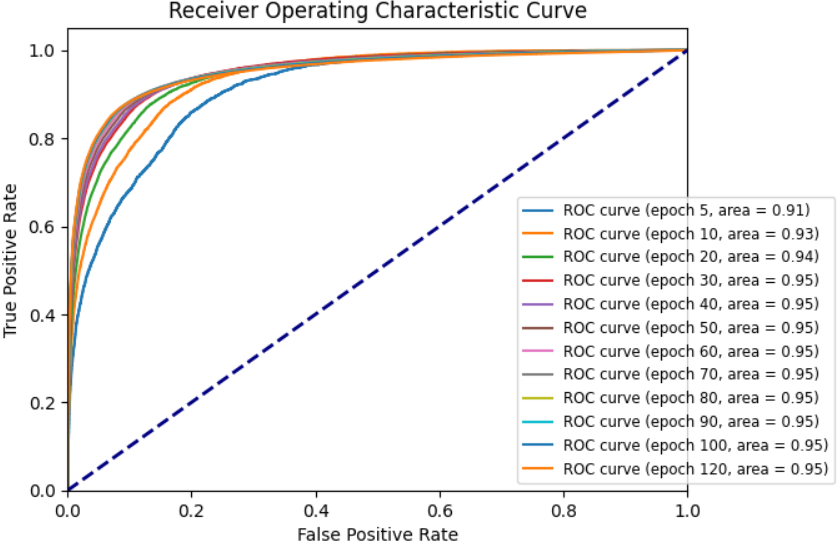}
    \caption{ROC curve for Ensemble model}
    \label{fig4}
\end{figure}

\vspace{-0.3cm}
The results of our experiments are presented in Table \ref{tab6}. As expected, the deeper DenseNet201 model surpassed the baseline C-Tran model that used DenseNet161. The ensemble models with deeper CNN backbones also demonstrated improved scores. The Strong Densenet and Weak Resnet ensemble variant 1, in particular, consistently achieved a model score above 0.91, surpassing the rest. It surpassed the baseline in 5 out of 8 metrics, scoring AUC less than 0.9 only in 1 label compared to 2 of the baseline. Fig. \ref{fig4} shows its ROC curve. As seen in Table 6, this model also surpassed the state-of-the-art (SOTA) model, which had a score of 0.9. As for the ViT-based models, all of its variants showed superior performance compared to the C-Tran baseline model while maintaining lower computational complexity. Notably, the IEViT v2 DPD variant displayed slightly better performance among the IEViT models for $32\times 32$ dimension patch. The IE variants of C-Tran also exhibited slight improvements over the baseline, showcasing that the IE concept can be advantageous even in generic transformer architectures while maintaining lower complexity.
\vspace{-0.6cm}

\vspace{-0.3cm}
\begin{table}[H]
\centering
\caption{ Comparative performance of the proposed hybrid models with the state-of-the-art (SOTA) model}
\resizebox{12cm}{3.5cm}{
\begin{tabular}{|c|c|c|c|c|c|}
\hline
\textbf{Model} & \textbf{ML mAP} & \textbf{ML F1} & \textbf{ML AUC} & \textbf{ML Score} & \textbf{Model Score} \\
\hline
\textbf{SOTA} & 0.573 & 0.685 & 0.962 & 0.824  & 0.9 \cite{b10} \\\hline
\textbf{Proposed Models} & \multicolumn{4}{c}{}& \\\hline
C-Tran DenseNet201  & 0.6909 & 0.6394 & 0.9221 & 0.8065  & \textbf{0.9032}$^{\mathrm{a}}$\\\hline
C-Tran ResNet152d  & 0.6580 & 0.6042 & 0.9315 & 0.7951 & 0.8973 \\\hline
EV1 (S: RN, W: DN) $^{\mathrm{b}}$   & 0.6686 & 0.5974 & 0.9279 & 0.7982  & 0.8991 \\\hline
EV1 (S: DN, W: RN)  & 0.7120 & 0.6831 & 0.9544 & 0.8332  & \textbf{0.9166} \\\hline
EV2 &  0.6904  & 0.6538 & 0.9303 & 0.8103 & \textbf{0.9052}
\\\hline
IEViT  &  0.6713 & 0.6015 & 0.9532 & 0.8122 & \textbf{0.9061} \\ \hline
IEViT v1 UP  & 0.6781 & 0.6070 & 0.9509 & 0.8145  & \textbf{0.9073}\\\hline
IEViT v2 DPD & 0.6893 & 0.6258 & 0.9420 & 0.8157 & \textbf{0.9078}\\ \hline
IECTe  &  0.6801 & 0.5993 & 0.9420 & 0.8111 & \textbf{0.9055} \\ \hline
IeECT  & 0.6725 & 0.6365 & 0.9299 & 0.8012 &  \textbf{0.9006}\\ 
\hline
\multicolumn{6}{l}{$^{\mathrm{a}}$Bold numbers show improved model scores $(>0.9)$}\\
\multicolumn{6}{l}{$^{\mathrm{b}}$S means strong, W means weak}
\end{tabular}}%
\label{tab6}
\end{table}

\pagebreak
\section{Conclusion}

 In summary, our research underscores the effectiveness of hybrid models in classifying retinal images despite the limitations of a small, imbalanced dataset. While we have made commendable progress in multi-label retinal disease classification, we acknowledge the constraints imposed by data size and class imbalances, which could lead to randomness and potential overfitting.

Despite these challenges, our work represents a significant step in advancing deep learning-based strategies for accurate retinal disease classification. It also opens avenues for further refinement. We plan to focus on improving model interpretability to address these limitations and enhance practical utility. In our future work, we aim to explore the integration of Shapley values to unravel the causal relationships behind our model's predictions, making them more transparent and understandable.

In conclusion, our study highlights the potential of combining transformers and ensemble learning to tackle retinal disease classification challenges. Through our ongoing efforts to enhance interpretability and delve into advanced techniques, we aim to pave the way for more robust, interpretable, and effective models that can significantly benefit the field of medical diagnostics.\\

%
%


\vspace{12pt}
\end{document}